\documentclass[11pt]{article} %
\usepackage{rldmsubmit,palatino}
\usepackage{graphicx,wrapfig}
\usepackage{tikz}
\usepackage{bm}
\usepackage[font=small]{caption}

\usepackage{layouts} %

\title{Prototyping three key properties of specific curiosity in computational reinforcement learning}

\author{
Nadia M. Ady\thanks{Corresponding author. Website: \href{https://ualberta.ca/\~nmady/}{https://ualberta.ca/$\sim$nmady/}} \\
Department of Computing Science\\
University of Alberta\\
Edmonton, AB T6G 2E8 \\
Alberta Machine Intelligence Institute \\
\texttt{nmady@ualberta.ca} \\
\And
Roshan Shariff \\
Department of Computing Science\\
University of Alberta\\
Alberta Machine Intelligence Institute\\
\texttt{roshan.shariff@ualberta.ca} \\
\AND
Johannes Günther \\
Alberta Machine Intelligence Institute \\
Department of Computing Science \\
University of Alberta \\
\texttt{johannes@amii.ca} \\
\And
Patrick M. Pilarski \\
Department of Medicine \\
Department of Computing Science \\
University of Alberta \\
Alberta Machine Intelligence Institute \\
\texttt{pilarski@ualberta.ca} \\
}

\begin{document}

\maketitle

\begin{abstract}

Curiosity for machine agents has been a focus of intense research. 
The study of human and animal curiosity, particularly \emph{specific curiosity}, has unearthed several properties that would offer important benefits for machine learners, but that have not yet been well-explored in machine intelligence. In this work, we introduce three of the most immediate of these properties---directedness, cessation when satisfied, and voluntary exposure---and show how they may be implemented together in a proof-of-concept reinforcement learning agent; further, we demonstrate how the properties manifest in the behaviour of this agent in a simple non-episodic grid-world environment that includes curiosity-inducing locations and induced targets of curiosity. As we would hope, the agent exhibits short-term directed behaviour while updating long-term preferences to adaptively seek out curiosity-inducing situations. This work therefore presents a novel view into how specific curiosity operates and in the future might be integrated into the behaviour of goal-seeking, decision-making agents in complex environments.

\end{abstract}


\acknowledgements{Thanks to all the amazing people who made this work better and clearer, especially Kate Pratt. The authors wish to thank their funding providers. NMA was supported by a scholarship from The Natural Sciences and Engineering Research Council of Canada (NSERC). Work by PMP was supported by grants or awards from the Canada CIFAR AI Chairs Program, the Alberta Machine Intelligence Institute (Amii), NSERC, and the Canada Research Chairs program.}

\startmain %

\section{Specific Curiosity: An Introduction}

Curiosity is in vogue; it is considered socially desirable, relating to a set of behaviours that employers want in their employees, teachers want in their students, and life coaches encourage in our interpersonal relationships. It should come as no surprise that machine intelligence researchers have wanted to provide the benefits of curiosity to their computational creations for decades (Schmidhuber, 1991). Pursuing these benefits, researchers have developed numerous mechanisms. 

\begin{wrapfigure}[15]{r}{0.3\textwidth}
\vspace{-1\baselineskip}
\includegraphics[width=0.3\textwidth]{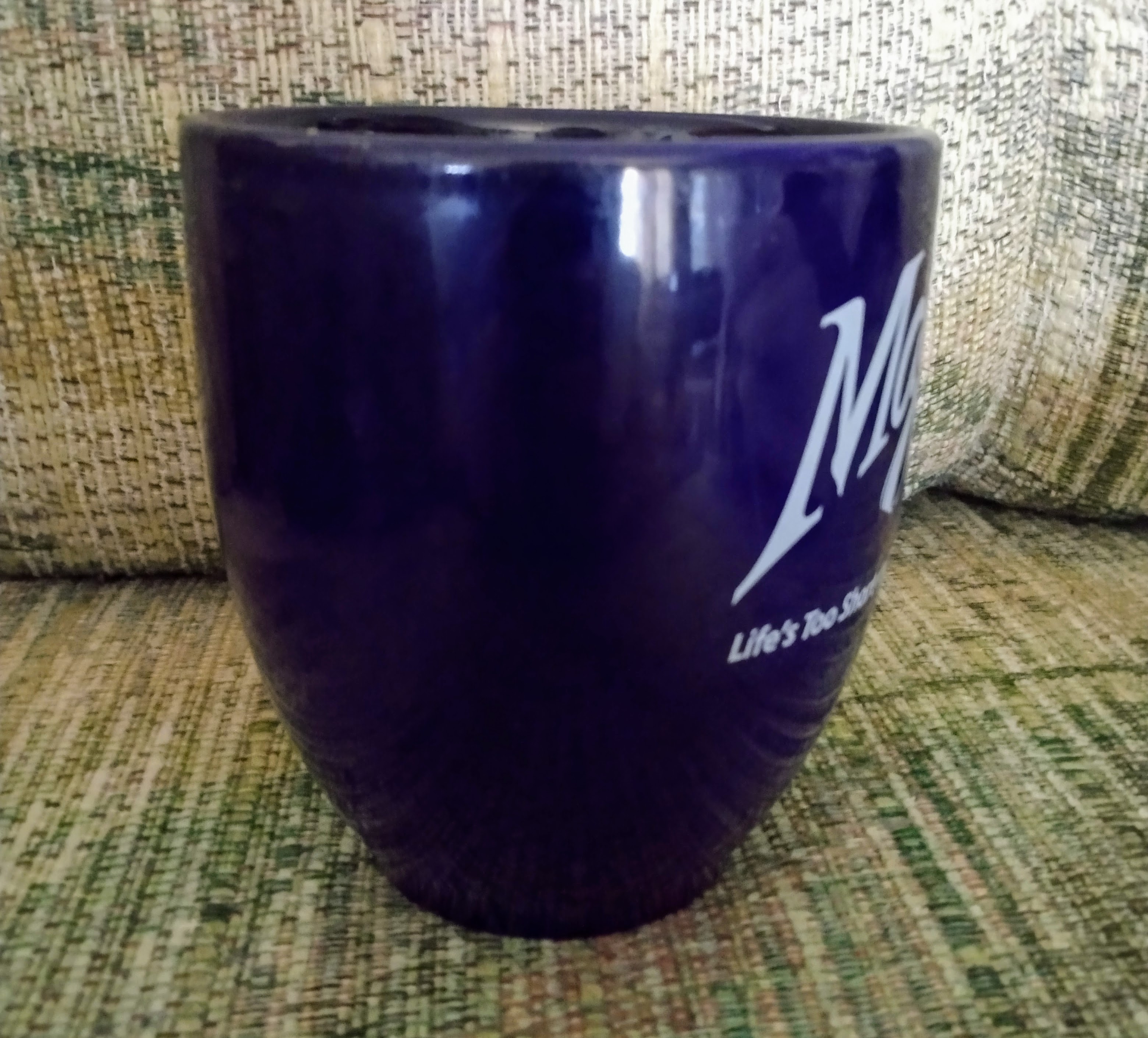}
  \caption{\label{fig:mug-inostensible}Turn to pg.~\ref{fig:mug-ostensible} to rotate the mug.} 
\end{wrapfigure}

While these mechanisms are different from each other (Ady \& Pilarski, 2017; Linke \textit{et al.}, 2019), many of them are centred on generating and using special reward-like signals called intrinsic rewards. 
Put simply, the learner, or agent, obtains intrinsic reward for encountering situations with good potential for learning.
Researchers have associated different concepts with ``potential for learning" and translated these concepts into metrics to shape the behaviour of their agents.\footnote{
Examples include confidence (Schmidhuber, 1991), learning progress (Oudeyer \textit{et al.}, 2007, p.~269), surprise (White \textit{et al.}, 2014, p.~14), interest/interestingness (Gregor \& Spalek, 2014, p.~435; Frank \textit{et al.}, 2014, pp.~5-6), novelty (Gregor \& Spalek, 2014, p.~435; Singh \textit{et al.}, 2004, pp. 1, 5), uncertainty (Pathak \textit{et al.}, 2017, pp.~1-2), compression progress (Graziano \textit{et al.}, 2011, p.~44), competence (Oddi \textit{et al.}, 2020, pp.~2417-2418), and information gain (Bellemare \textit{et al.}, 2016, p.~4; Houthooft \textit{et al.}, 2016, pp.~ 2-3).}
In general, these metrics are used to determine the amount of intrinsic reward the agent receives for a given experience. Intrinsic rewards have been shown to be very useful for increasing exploration in some situations (Pathak \textit{et al.}, 2017). 

However, intrinsic rewards produce behaviour that is conceptually unlike curiosity. Critically, by rewarding a learner for getting into a particular situation, the designer is encouraging the learner to return to that same situation, or ones very like it. This behaviour is missing a conceptual aspect of curiosity.

When humans describe curiosity, they talk about their curiosity being piqued or triggered. They take action, not to observe the same situation that piqued their curiosity, but to observe a different situation that they believe will satisfy their curiosity (see Fig.~\ref{fig:mug-inostensible}). Humans take advantage of their learned knowledge of how the world works to learn something specific, answering a question that the curiosity-piquing situation led them to ask. This is sometimes referred to as {\em specific curiosity}.

As a main contribution, we distill from the related literature three key properties of specific curiosity, and argue that these properties are beneficial to machine learners; specifically, we introduce the appropriateness of a reinforcement learning approach for designing agents that exhibit these properties. These properties are as follows, and are a subset of five key properties described and argued for in more detail by Ady \textit{et al.} (2022).

\begin{description}
\item[Directedness:] By separating a curiosity-piquing situation from a curiosity-satisfying situation, we can think about the appropriate behaviour for a learner whose curiosity has been piqued: take a sequence of actions that seems most likely to take them to a situation that satisfies their curiosity. The tendency for learners experiencing specific curiosity to undertake directed behaviour has been well-documented in the literature (e.g. Hagtvedt \textit{et al.} 2019, p.~2; Berlyne, 1960, p.~297). This view contrasts with many other machine reinforcement learning methods which rely on injecting randomness into their choices of actions to experience new situations, rather than heading directly for a situation they know they don't know.
\item[Cessation when satisfied:] Once a learner has found a situation that satisfies their curiosity, they don't need to experience the same situation again, so we have no need to incentivize returning to a curiosity-satisfying situation. Examples documenting the satisfiability of curiosity are prevalent in the literature, including the works of Wiggin \textit{et al.} (2019, p.~1194), Buyalskaya \& Camerer (2020, p.~141, who refer to it as `fulfillment') and Dan \textit{et al.} (2020, p.~150, who refer to curiosity being `satiated').
\item[Voluntary exposure:] While the learner does not benefit from returning to a curiosity-satisfying situation (that would only answer a question for which they have a satisfactory answer!), they do benefit from experiencing curiosity-inducing situations. 
A curiosity-inducing situation offers an appropriate jumping-off point for learning, for metaphorically picking up a puzzle piece fitting into what the learner already knows.
Choosing to partake in activities likely to induce curiosity---for example, picking up puzzles or mysteries or turning on Netflix---has been called voluntary exposure to curiosity (Loewenstein, 1994, p.~84). This property might be thought of as developing an increased preference for situations like those that have been curiosity-inducing in the past.
\end{description}

Finally, we provide a case study of an agent demonstrating these three properties together, actively separating curiosity-piquing situations from curiosity-satisfying situations---a separation which does not exist in the type of behaviour motivated by intrinsic rewards. This proof-of-concept shows that it is possible to create an agent with attributes of specific curiosity, but should not be thought of as \emph{the} way to implement specific curiosity. Rather, we hope it inspires the community to build upon our ideas and think up improved machine agents that benefit from knowledge of specific curiosity.

\section{The Appropriateness of a Reinforcement Learning Framework for Specific Curiosity}
\label{sec:rlsc}

\begin{wrapfigure}[20]{r}{0.3\textwidth}%
\vspace{-1\baselineskip}%
\includegraphics[width=0.3\textwidth]{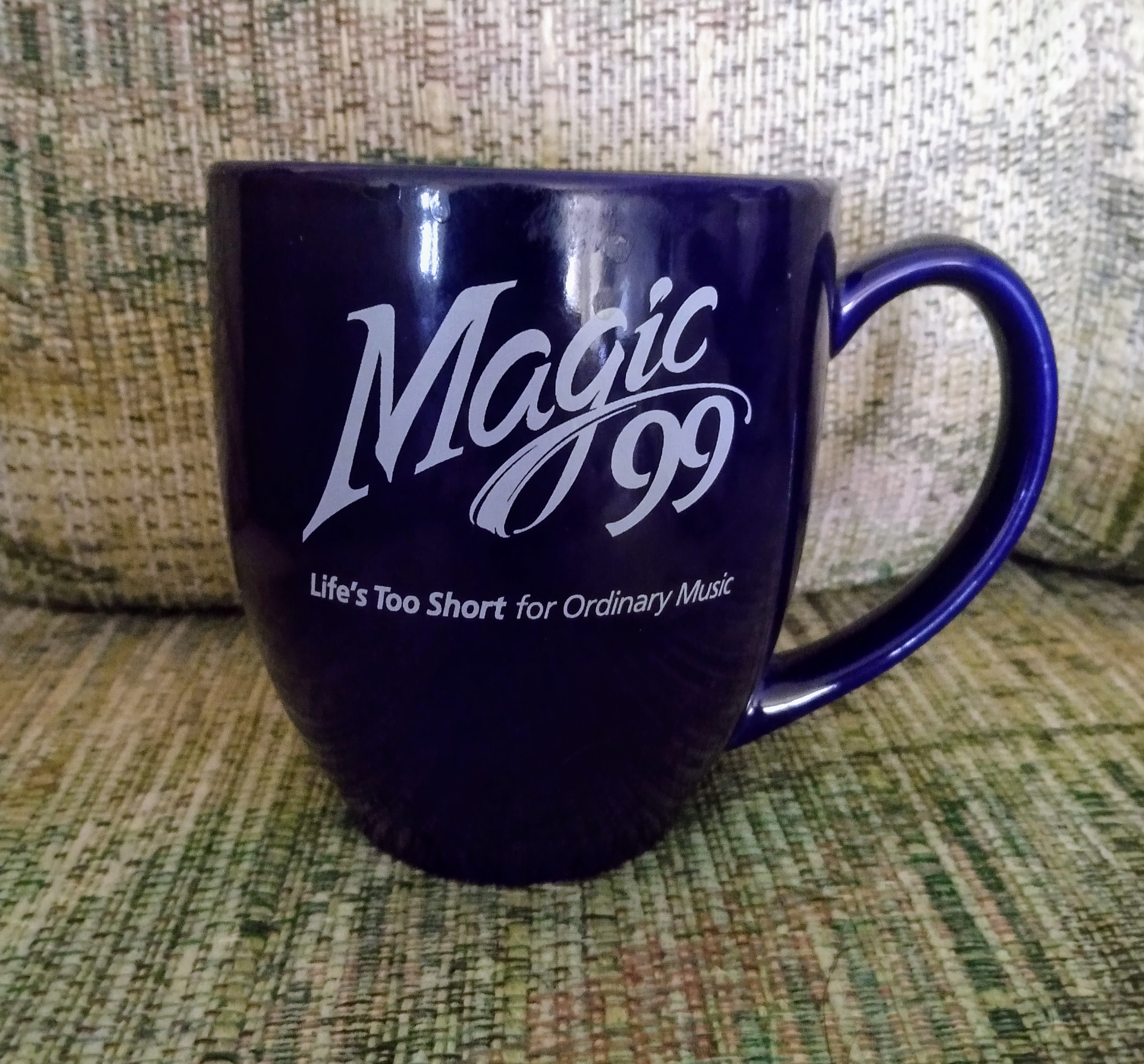}
  \caption{\label{fig:mug-ostensible} If you were curious about what was written on the mug, you probably didn't keep staring at Fig.~\ref{fig:mug-inostensible}. Instead, you took action to reach a situation that would satisfy your curiosity.}%
\end{wrapfigure}

With the goal to offer the benefits of specific curiosity to a computational learner, we want to take advantage of an existing learning framework, ideally one that supports the properties we want to achieve. The properties of directedness and voluntary exposure require a learner to have some specific capabilities. For directedness, the learner must be able to decide how to act in the world so they can take the specific actions that they believe will allow them to satisfy their curiosity. For voluntary exposure, the learner must be able to decide how to act in the world so they can visit situations that they suspect will result in their curiosity being piqued, and also learn those enduring preferences for curiosity-inducing situations.

The framework of reinforcement learning supports these requirements, as reinforcement learning centres around learners %
who are able to choose actions that affect their experiences (Sutton and Barto, 2018, p.~3). Instantiations of computational reinforcement learning algorithms are often called agents because they shape their own experience in the world and learn from their actions. This quality makes the framework well-suited for the design of machine curiosity algorithms.

A reinforcement learning framework is particularly promising for the property of voluntary exposure, as learners can estimate the value of different situations (for more detail, see Sutton \& Barto, 2018, p.~58). In this way, a learner may develop a preference (i.e., a  higher value) for situations that are more likely to pique curiosity. This preference aligns with voluntary exposure as observed in humans. 

While, to the best of our knowledge, our conceptual separation of curiosity-piquing situations from curiosity-satisfying situations and our argument for specific properties of specific curiosity are new contributions, computational reinforcement learning researchers have shown strong interest in aspects of these properties in other contexts. In particular, the property of directedness parallels work done on options (as early as Sutton \textit{et al.}, 1999) and planning. Approaches using directedness also often exhibit cessation when satisfied. For example, the options framework includes termination conditions for each option (often naturally defined by goal states; Stolle \& Precup, 2002, p.~212).

Prior work has aimed to address the lack of directedness that is a characteristic of intrinsic-reward methods. For example, the Model-Based Active eXploration algorithm presented by Shyam \textit{et al.}\ (2019) focuses on planning behaviour to allow the agent ``to observe novel events" (p.~1). Similarly, the Go-Explore family of algorithms centres on the idea of taking a direct sequence of actions to move to a specific state for the purpose of exploring from it (Ecoffet \textit{et al.}, 2020).

\section{Case Study \& Experiments}
In this section, we describe how we created a simple agent exhibiting some of the properties of specific curiosity. We specifically hope to show that, even in a simple and focused setting, using our properties as guidelines allows machine behaviour to emerge that approximates the specific curiosity of animal learners. %
This example is not intended as
a final or definitive computational implementation of specific curiosity. The intended purpose of this section is for the reader to gain insight and motivation to further investigate how to integrate the properties of specific curiosity into different machine learning frameworks and problem settings.

\begin{wrapfigure}[44]{r}{0.3\textwidth}%
\vspace{-1\baselineskip}%
\includegraphics[trim=0.25cm 0.5cm 0.25cm 0.4cm]{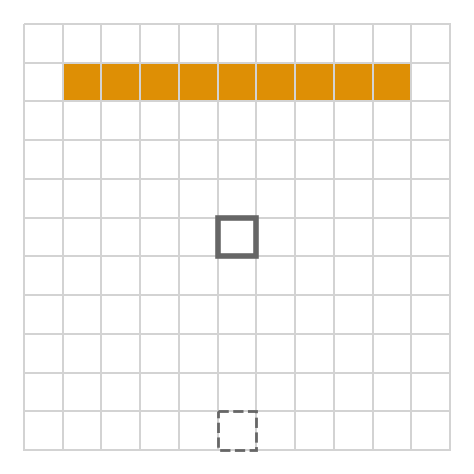}
  \caption{\label{fig:domain} {Domain used for the experiments.} The orange boxes are the potential curiosity-satisfying locations. The heavily-outlined box is the curiosity-inducing location. If the agent leaves the top row, it teleports to the dashed box in the bottom row.} 
  
  \vspace{3em}
  
  \includegraphics[trim=0.25cm 0.5cm 0.25cm 0.4cm]{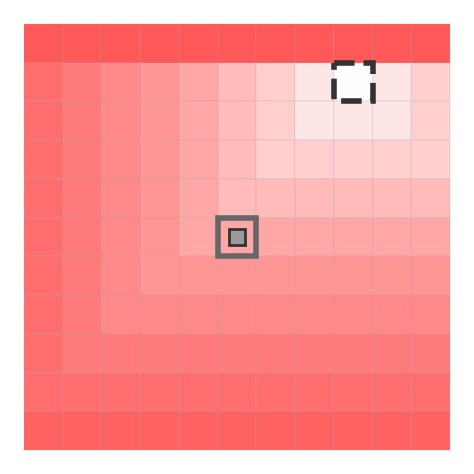}
  \caption{\label{fig:vcurious} {An example temporary value function}, $V_\mathrm{curious}$. The target has a heavy dashed outline and a value of zero. Magnitudes of negative values are shown using shades of red.}
\end{wrapfigure}

A complete agent with specific curiosity will require some additional functionality that we have not implemented.
A human can recognize that there is something they do not know, sometimes referred to as an \textit{information gap} (Loewenstein, 1994) or an \textit{inostensible concept} (Inan, 2012). Moreover, humans can imagine what they would need to observe to rectify such a gap, and even suggest actions that might lead to those observations. For example, you could reach out and turn the mug in Fig.~\ref{fig:mug-inostensible}, or move your own body to see the mug at a different angle. We do not tackle how agents can recognize information gaps or how they can predict what will satisfy their resulting curiosity. In our case study, we gave the agent a module that could recognize a curiosity-inducing situation and a corresponding curiosity-satisfying target. We hard-coded this module so only a single location in the world piqued curiosity and the targets were randomly selected locations from a pre-determined subset (see Fig.~\ref{fig:domain}). We envision that the gaps in this architecture can be filled as more aspects of specific curiosity are understood and become computationally tractable.

\textbf{Directedness: }The first property we explored in the design of our agent was directedness. As we have already suggested, directedness seems to involve making and following a plan, which, in a reinforcement learning framework, suggested to us that we might use a model of the world. In computational reinforcement learning, a model-based approach using dynamic programming or an approximation of dynamic programming can provide a learner with a plan to follow from one state to another (Sutton \& Barto, 2018, Ch.~4), suggesting existing algorithms we could exploit in our pursuit of the property of directedness.

\textbf{Cessation when satisfied: }
Directedness in the pursuit of satisfying curiosity is temporary and should not persist when the agent is not in a state of curiosity.\footnote{In the literature, there is a distinction between curiosity as a \textit{state} experienced temporarily by a learner versus as a \textit{trait}, or general propensity of the learner (Loewenstein, 1994, p.~78). Specific curiosity makes the most sense as a form of \textit{state} curiosity because it only persists until the specific information of interest is found.}
We were inspired by an architecture introduced by Silver \textit{et al.} in 2008: Dyna-2.
We realized that Dyna-2 has a mechanism (originally used for playing games) that differentiates temporary behaviour from the long-term preferences of the agent: two value functions. %
In Silver's (2009, p.~87) work, a temporary value function captures unique aspects of the current match of a game, while a persistent value function captures principles of the game that hold across matches. In our design, we adapt this idea so the learner follows the temporary value function while curious but uses a separate persistent value function to capture long-term preferences. The temporary value function can lead the learner directly to a situation they believe will satisfy their curiosity. Since what will satisfy curiosity differs each time curiosity is piqued, the exact plan to satisfy curiosity can be discarded at satisfaction.

\textbf{Voluntary exposure: }A temporary value function doesn't have a lasting effect on the learner's preferences. However, we want an enduring effect associated with curiosity: voluntary exposure. Humans develop a preference for curiosity-inducing situations, so we wanted to experiment with algorithm modifications that would lead to agents with this property. We came up with a small modification to the standard TD learning algorithm (Sutton \& Barto, 2018, p.~120) that would leave an enduring effect of having experienced curiosity on the persistent value function, $V$. Our modified temporal difference, $\delta$, is:
\begin{equation}
    \delta \gets R + \gamma \cdot V(x') - \color{blue}\bm{[ V(x) + V_\mathrm{curious}(x)]} \label{eq:td-update}
\end{equation}
where $V_\mathrm{curious}$ refers to the temporary value function. %
For this modification to result in the accrual of positive value in the persistent value function, $V$, it is necessary for $V_\mathrm{curious}$ to be non-positive everywhere. Given this modification, we ran experiments to see if the enduring effect looked like voluntary exposure.

\textbf{Experiment setup: } We ran our experiments in an $11 \times 11$ grid world shown in Fig.~\ref{fig:domain}. In this grid world, the learner is located in one of the squares and their actions let them move to adjacent squares, but they can only take sideward or upward actions (including diagonals). Any upward action from the top row of the grid teleports the agent to the middle of the bottom row. The agent never receives any reward from this domain ($R$ in Eq. \ref{eq:td-update} is always zero). 

As long as the agent is not in a state of curiosity, the temporary value function is zero and the agent acts $\epsilon$-greedily with respect to its persistent value function. The centre of the grid is a curiosity-inducing location: when the agent visits it, a curiosity-satisfying target is generated and the agent is given a non-positive temporary value function (Fig.~\ref{fig:vcurious}) that guides it directly to that target. The target is randomly selected with equal probability for each of the non-edge squares in the second row from the top (highlighted in Fig. \ref{fig:domain}). When the agent visits the target, the temporary value function is zeroed out again.

The results reported here reflect 30 trials of 5000 steps. The agent started each trial in the curiosity-inducing location. We ran further experiments varying the size and dynamics of the grid world, and the key results we present are representative of all these experiments.

\begin{wrapfigure}[40]{r}{0.3\textwidth}
  \vspace{-1\baselineskip}
  \includegraphics[trim=0.25cm 0.25cm 0.4cm 0.25cm]{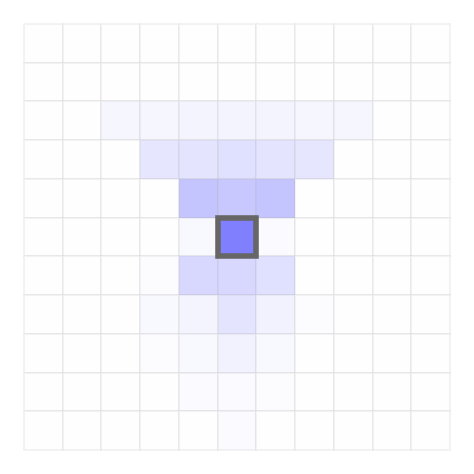}
  \caption{\label{fig:value} {Persistent value function}, $V$, learned by the end of 5000 steps, averaged over 30 trials. White squares have zero value. More positive values are shown in deeper shades of blue.}
  
  \begin{tikzpicture}
  \node[anchor=north] (domain) at (0,0) {\includegraphics[trim=0.2cm 0.4cm 0.2cm 0, clip]{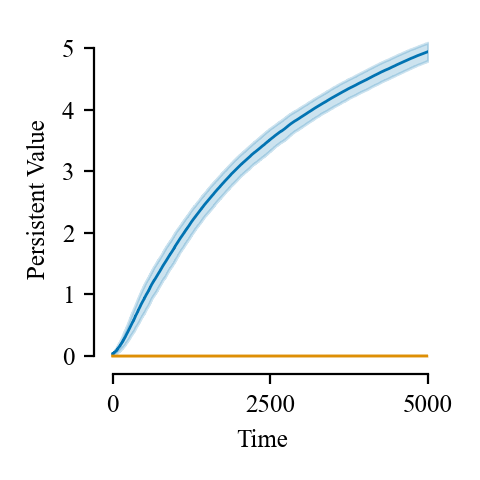}};
  \node[align=center, anchor=south] (inducer_label) at (-0.1,-1.5) {Curiosity-inducing\\location};
  \node (inducer_endpoint) at (0.1,-2) {};
  \draw[->] (inducer_label) to [out=270,in=130] (inducer_endpoint);
  \node[align=center, anchor=north] (satisfier_label) at (0.8,-3.6) {Curiosity-satisfying\\locations};
  \node[align=center, anchor=north] (satisfier_forepoint) at (1.5,-4.1) {};
  \node (satisfier_endpoint) at (1.9,-4.6) {};
  \draw[->] (satisfier_forepoint) to [bend left] (satisfier_endpoint);
  \end{tikzpicture}
   \caption{\label{fig:lineplot} Persistent value of the curiosity-inducing location is in blue while the average persistent value of all of the curiosity-satisfying locations is in orange. The lines are the average over 30 trials while the shaded area shows the standard deviation.} 
\end{wrapfigure}

\textbf{Results \& Discussion: }Fig.~\ref{fig:value} shows the persistent value function at the end of 5000 steps, averaged over 30 trials. Fig.~\ref{fig:lineplot} shows how the values of the curiosity-inducing location and the curiosity-satisfying locations change over time.

The curiosity-inducing location accrues the most value and the agent learns a trail of increasing value leading directly to that location. The curiosity-satisfying locations, on the other hand, do \textit{not} accumulate much value over time. In effect, our modification to the learning update results in voluntary exposure while retaining the property of cessation when satisfied.

Looking at the behaviour of the agent over time (an example video can be viewed at \url{https://youtu.be/TDUpB7OefFc}) we can see the agent moves more and more directly towards the curiosity-inducing location.

\section{Conclusions}

We argue that learners benefit from specific curiosity, something previously not demonstrated in the computational literature.
In particular, specific curiosity has properties of directedness, cessation when satisfied, and voluntary exposure that yield important benefits that have been missing from other approaches to machine curiosity.
In this work we developed a case study showing how these properties might be implemented in a computational reinforcement learning framework.

We appreciate reinforcement learning for its flexibility to express key properties of specific curiosity. 
We found, through our case study, that separating persistent and temporary value functions was a useful mechanism to encode some of the properties of biological curiosity. This separation gives our agent directed behaviour as well as learned preferences, which have traditionally been associated with model-based and model-free approaches to reinforcement learning, respectively. The way our agent balances these two approaches may offer a new perspective on how we might balance model-based and model-free learning.

\textbf{Future directions: } %
While directedness, cessation when satisfied, and voluntary exposure are the first properties we have explored empirically, our conceptual synthesis of the literature also demonstrates the importance of two more properties: transience and connection to long-term information search (Ady \textit{et al.}, 2022). While our agent has begun to exhibit some aspects of specific curiosity, including these additional properties and continuing to incorporate new ideas from the study of human curiosity can only strengthen our computationally curious agents.

\section*{References}
\setlength{\parskip}{0.25\baselineskip}
\footnotesize

\noindent Ady, Shariff, G\"{u}nther, and Pilarski, in preparation for \textit{J. Artificial Intelligence Research}, 2022.

\leftskip 0.25in
\parindent -0.25in

Ady and Pilarski, {\em Multidiscip.~Conf.~Reinforcement Learning and Decision Making}, 2017.

Bellemare, Srinivasan, Ostrovski, {\em et al.}, {\em NeurIPS}, 2016, 1471--1479.

Berlyne, {\em Conflict, arousal, and curiosity.} McGraw-Hill Book Company, 1960.

Buyalskaya and Camerer, {\em Current Opinion in Behavioral Sciences}, 2020, 35:141--149.

Dan, Leshkowitz, and Hassin, {\em Current Opinion in Behavioral Sciences}, 2020, 35:150--156.

Ecoffet, Huizinga, Lehman, Stanley, and Clune, {\em Nature}, 2021, \textit{590}, 580--586.

Frank, Leitner, Stollenga, {\em et al.}, {\em Frontiers in Neurorobotics}, 2014, \textit{7}:25.

Graziano, Glasmachers, Schaul, {\em et al.}, {\em Acta Futura}, 2011, \textit{4}, 41--52.

Gregor and Spalek, {\em ELEKTRO 2014 Conference}, 2014, 435--440.

Hagtvedt, Dossinger, Harrison, {\em et al.}, {\em 
Org.~Behav.~and Human Decision Proc.}, 2019, 150:1-13.

  Houthooft, Chen, Duan, {\em et al.}, {\em NeurIPS}, 2016, 1109--1117.

  Inan, {\em The Philosophy of Curiosity}, 2012. Routledge.

  Linke, Ady, Degris, {\em et al.}, {\em Multidiscip.~Conf.~Reinforcement Learning and Decision Making}, 2017.

  Loewenstein, {\em Psychological Bulletin}, 1994, \textit{116}:1, 75--98.

  Markey and Loewenstein, {\em International handbook of emotions in education}, 2014, 238--255.

  Pathak, Agrawal, Efros, {\em et al.}, {\em Int. Conf. on Machine Learning}, 2017, 2778--2787.

  Oddi, Rasconi, Santucci, {\em et al.},  {\em European  Conference  on  Artificial  Intelligence}, 2020, 2417--2424.

  Oudeyer, Kaplan, and Hafner, {\em IEEE Trans. on Evolutionary Computation}, 2007, \textit{11}:2, 265--286.

  Schmidhuber, {\em International Joint Conference on Neural Networks}, 1991, 1458--1463.

  Shyam, Ja\'{s}kowski, and Gomez, {\em Int. Conf. on Machine Learning}, 2019, 5779--5788.

  Silver, Sutton, and M\"{u}ller, {\em Int. Conf. on Machine Learning}, 2008, 968--975.

  Silver, ``Reinforcement Learning and Simulation-Based Search in Computer Go," Doctoral Thesis, 2009. 

  Singh, Barto, and Chentanez, {\em NeurIPS}, 2004.
  
  Spielberger and Starr, In: {\em Motivation: Theory and Research}, 1994, 221--243. Lawrence Erlbaum Associates, Hillsdale, New Jersey.

  Stolle and Precup, {\em  International Symposium on Abstraction, Reformulation, and Approximation}, 2002, 212--223.

  Sutton and Barto, {\em Reinforcement Learning: An Introduction}, 2018. MIT Press, second edition.

  White, Modayil, and Sutton, {\em Sequential Decision-Making with Big Data: AAAI-14 Workshop}, 2014, 19--23.
  
  Wiggin, Reimann, and Jain, {\em Journal of Consumer Research}, 2019, 45(6):1194--1212.

\end{document}